\documentclass{article}
\usepackage{spconf,amsmath,graphicx}

\usepackage{subfigure}

\usepackage[hidelinks]{hyperref}
\usepackage{xcolor}

\usepackage{ragged2e} 
\usepackage{booktabs,makecell, multirow, tabularx}

\usepackage{enumitem}

\title{Enabling Energy-Efficient Object Detection with Surrogate Gradient Descent in
	Spiking Neural Networks}
\name{Jilong Luo, Shanlin Xiao$^{\ast}$, Yinsheng Chen, Zhiyi Yu$^{\ast}$
	\thanks{*Corresponding author: Shanlin Xiao, xiaoshlin@mail.sysu.edu.cn; Zhiyi Yu, yuzhiyi@mail.sysu.edu.cn.} 
	\thanks{$^1$ The code is made available at \textcolor{blue}{\href{https://github.com/xiaolongren969/SNN-YOLOv3}{https://github.com/xiaolongren969/ SNN-YOLOv3}}}
	}

\address{Sun Yat-sen University, China}
\begin{document}
%
\maketitle
\begin{abstract}
	Spiking Neural Networks (SNNs) are a biologically plausible neural network model with significant advantages in both event-driven processing and spatio-temporal information processing, rendering SNNs an appealing choice for energy-efficient object detection. However, the non-differentiability of the biological neuronal dynamics model presents a challenge during the training of SNNs. Furthermore, a suitable decoding strategy for object detection in SNNs is currently lacking. In this study, we introduce the Current Mean Decoding (CMD) method, which solves the regression problem to facilitate the training of deep SNNs for object detection tasks. Based on the gradient surrogate and CMD, we propose the SNN-YOLOv3 model for object detection. Our experiments\\ demonstrate that SNN-YOLOv3 achieves a remarkable performance with an mAP of 61.87\%  on the PASCAL VOC dataset, requiring only 6 time steps. Compared to Spiking-YOLO, we have managed to increase mAP by nearly 10\% while reducing energy consumption by two orders of magnit-
	\\ude$^1$.
\end{abstract}
\section{Introduction}
\label{sec:intro}


As a third-generation artificial neural network \cite{maass1997networks}, Spiking Neural Networks (SNNs) are promising for implementing low-power artificial intelligence algorithms using event-driven neuromorphic hardware \cite{merolla2014million, pei2019towards, indiveri2015neuromorphic}. Based on biological plausibility \cite{mainen1995reliability}, SNNs emulate information processing mechanisms observed in the biological neural system, where computation and information transfer between neurons occur through discrete binary events \cite{roy2019towards}. Despite the attractive energy efficiency of spiking neural networks, training SNNs remains a significant challenge. One of the primary reasons for this challenge is the complexity of the dynamics model and the non-differentiability of spiking neurons, typically modeled as IF or LIF neurons, which makes performing gradient descent-based backpropagation difficult \cite{lee2016training, shrestha2018slayer}.

\begin{figure*}[tbp] 
	\centering 
	\includegraphics[width=0.95\textwidth]{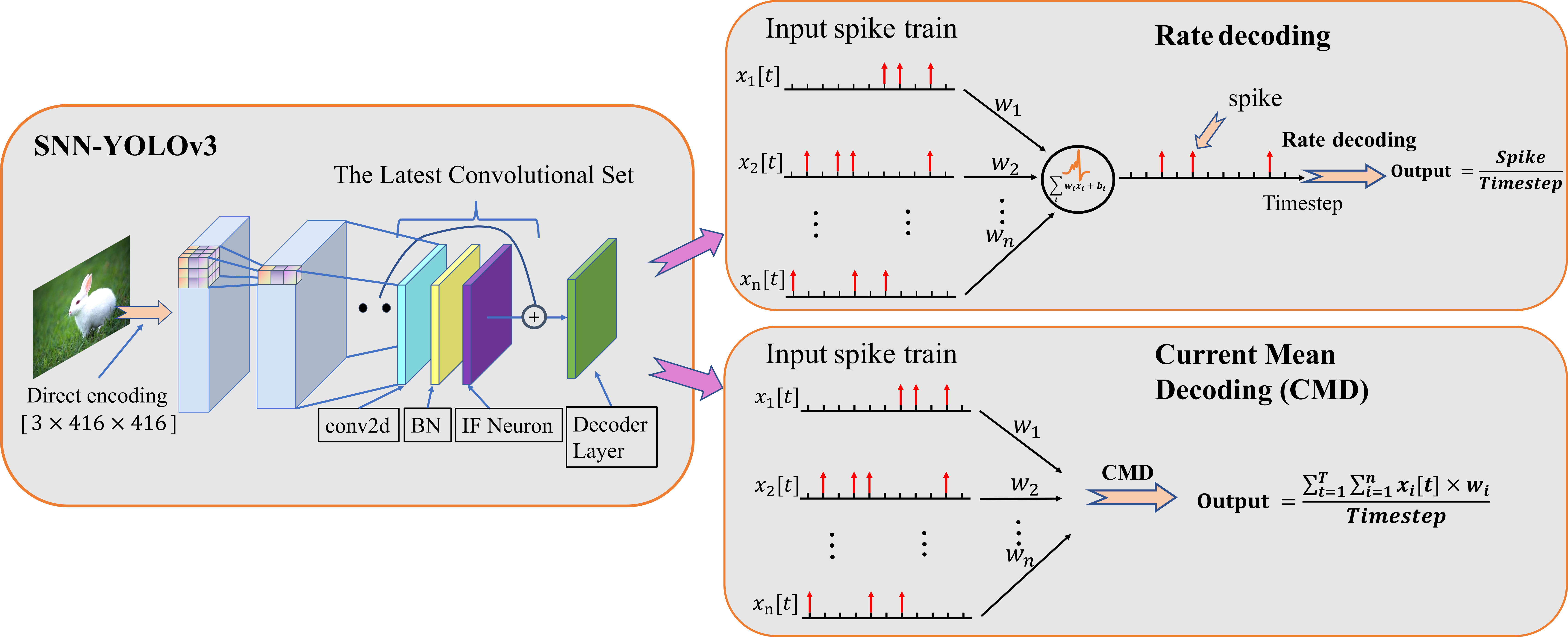} 
	\caption{Traditional rate decoding and proposed current mean decoding (CMD) in SNN-YOLOv3. The left part shows that the input of the image is coded by a direct encoding way and the activation layer in the network is replaced with IF neurons. The last layer is the decoding layer, and the right part shows the different decoding methods used in the decoding layer.} \label{CMD} 
\end{figure*}

Several researchers have proposed training SNNs using ANN-to-SNN conversion method \cite{diehl2015fast, sengupta2019going}, in which ANNs (Artificial Neural Networks) with ReLU activation function are initially trained via gradient descent and then converted into SNNs with integrate-and-fire neurons by applying appropriate threshold balancing techniques \cite{rueckauer2017conversion}. 
However, SNNs obtained through ANN-to-SNN methods generally occur 2000-3000 time steps to achieve acceptable accuracy. Here, a time step denotes the time unit for the forward propagation of a single layer, effectively represents network latency \cite{kim2020spiking}. 

To reduce the latency, the gradient surrogate-based backpropagation algorithm \cite{lee2016training, neftci2019surrogate} has been introduced for end-to-end gradient descent learning on spiking train. Within these algorithms, the non-differentiable neuron model completes the backpropagation process by specifying a surrogate gradient as a continuous approximation of the actual gradient \cite{wu2018spatio}. Training SNNs with gradient surrogate substantially  decreases the inference latency by nearly 100x (e.g., only requiring fewer than 30 time steps). Despite the appealing property of SNNs, previous research has mainly focused on less complex tasks (image classification) and small-scale datasets (MNIST and CIFAR10), with relatively shallow network structures (\textless 30 layers) \cite{kheradpisheh2020temporal, zhou2021temporal}.

In this study, we investigate more complex machine learning problems (object detection) in deep SNNs, using the gradient surrogate approach. Object detection is considered a demanding and challenging task in computer vision, aiming to recognize multiple objects and calculate the exact coordinates of the bounding box in images or videos. Unlike image classification tasks, when predicting the output values of a neural network, object detection requires predicting continuous numerical or real outputs, rather than just selecting the category with the highest probability (using the argmax function), as is typically done in image classification tasks. Our contributions can be summarized as follows:

\noindent
\begin{itemize}[leftmargin=*]
	\setlength{\itemsep}{5pt }
	\item We present the first SNN model that implement object detection using surrogate gradient to achieve the state-of-the-art performance (61.87\% mAP) on the non-trivial dataset of PASCAL VOC.
	\item We introduce the Current Mean Decoding (CMD) method, which solve the regression problem to facilitate the training of deep SNNs for object detection tasks.
\end{itemize}

\section{methods}
\subsection{Surrogate gradient for spiking neuron models }

Unlike ANNs, SNNs utilize spike trains for computation and information transmission among neurons. The dynamics of the classic IF neuron model \cite{cao2015spiking, han2020rmp} can be described as follows:

\begin{equation}  
	V_{mem,j}^l\ [t]=V_{mem,j}^l [t-1]+I_j^l [t]-V_{th} s_j^l [t]
\end{equation}
where  $s_j^l [t]$ represents the spike state of the j-th neuron in the l-th layer at time step t. $x_j^l [t]$ represents the input membrane potential, and $V_{mem,j}^l [t]$ represents the membrane potential of the j-th neuron in the l-th layer. The description of $I_j^l [t]$ can be expressed as follows:

\begin{equation}  
	I_{j}^{l}\left[ t \right] =\sum_i{w_{i,j}^{l}s_{i}^{l-1}\left[ t \right] +b_{j}^{l}}
\end{equation}
where $w$ and $b$ represent the synaptic weights and bias, respectively. When the membrane potential $V_{mem,j}^l [t]$ of the j-th neuron in the l-th layer exceeds the threshold voltage $V_{th}$, a spike $s_j^l [t]$ is emitted. The mathematical formula is as follows:

\begin{equation}  
	s_{j}^{l}\left[ t \right] =H\left( V_{mem,j}^{l}\left[ t \right] -V_{th} \right) 
\end{equation}
the mathematical description of $H(\cdot)$ is the Heaviside step function. This function produces a value of 1 when x is greater than or equal to 0, and 0 otherwise. Due to the non-differentiability of Heaviside unit step function, the surrogate gradient method is used to estimate gradient computations during backpropagation. The fundamental idea of surrogate gradient is to update the weights using gradient backpropagation via a surrogate gradient function rather than the unit step function. In this research, we have chosen the arctangent function as our surrogate gradient function. Its mathematical expression is illustrated below:

\begin{equation}  
	g\left( x \right) =\ \frac{1}{\pi}\arctan \left( \frac{\pi}{2}\alpha x \right) +\frac{1}{2}
\end{equation}
 Its derivative is represented as follows:
\begin{equation}  
	g^{'}\left( x \right) =\frac{\alpha}{2\left( 1+\left( \frac{\pi}{2}\alpha x \right) ^2 \right)}
\end{equation}


\subsection{Current mean decoding (CMD)}
Rate decoding is a commonly used approach to transder information in spiking neural networks that decodes information intensity dependent on the rate of neuron spike emissions \cite{auge2021survey, diehl2015unsupervised}. Nonetheless, in specific tasks, particularly those concerned with regression problems like object detection, rate decoding will struggle and require more advanced decoding strategies. 

Object detection  generally involves two primary tasks: classification and regression. In the classification task, the classification result can be determined by the max magnitude of the output neuron spike firing rate (rate decoding) in SNNs. However, it is often necessary to predict the location, size and shape of the object in regression tasks, which requires the output value space of network to be real-valued. Nevertheless, the discrete spikes employed for rate decoding are not directly mapped to a continuous numerical space, resulting in the network output being discrete. Consequently, the discrete event of rate decoding might lead to a loss of accuracy when representing continuous outputs. While continuous values may be approximated through rate decoding, this approximation need a compromise between precision and time step.

For this reason, we introduce current mean decoding(here-after abbreviated as CMD) in spiking neural networks, which is a more powerful decoding technique that exploits the dynamic properties of neurons. Fig.\ref{CMD} illustrates a schematic diagram of CMD, which collects the currents produced at synapses upon neuron spike event. Subsequently, it accomplishes information decoding by computing the mean value of input current, which is defined as follows.

\begin{equation}  
	Output=\frac{\sum\limits_{t=1}^T{\sum\limits_{i=1}^n{x_i\left[ t \right] \times w_i}}}{Time step}
\end{equation}

where $x_i[t]$ represents the spike firing spike state of the $i$-th neuron of the presynaptic neuron at time step $t$, and $w_i$ represents the weight associated with the corresponding neuron. These values are multiplied to yield the synaptic current.

On the comparison with rate decoding, this decoding method provides a better approximation of continuous values and offers better accuracy and flexibility for regression problems.


\begin{figure}[tbp] 
	\centering 
	\includegraphics[width=0.5\textwidth]{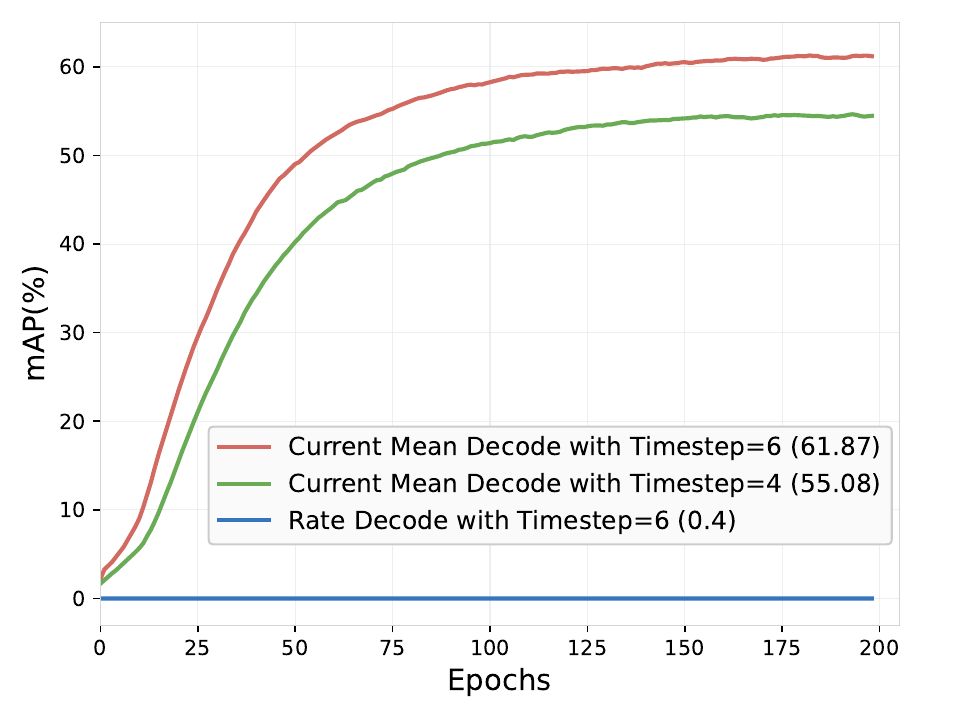} 
	\caption{Experimental results of SNN-YOLOv3 on PASCAL VOC dataset for various time step; maximum mAP is in parentheses.} \label{map}
\end{figure}

\section{EXPERIMENTS}
\subsection{Experimental setup}
In this experiment, we choose the classical version of the YOLOv3 \cite{redmon2018yolov3} real-time object detection network for validating the effectiveness of the CMD method. SNN-YOLOv3 was tested on the PASCAL VOC datasets \cite{everingham2010pascal} with simulations based on the PyTorch platform and all experiments were done on NVIDIA Tesla V100 GPUs. The SNN-YOLOv3 obtains more efficiency and robust spike feature trains by using the first convolutional set as the encoding layer without additional encoding layers.

During the training process, we adopt a stochastic gradient descent optimiser with a momentum parameter of 0.9 and a cosine decay scheduler for fine-tuning the learning rate. we set the weight decay for the upper bound parameter $\theta$ of SNN-YOLOv3 to $5\times10^{-3}$. Furthermore, We used normalization and horizontal flipping method for data augmentation.

\subsection{Experimental results}

In order to verify and analyse the effectiveness of our proposed method, we evaluate the performance of our methods for object detection tasks on PASCAL VOC datasets.

\begin{figure}[tb]
	\centering
	\subfigure{\includegraphics[width=0.23\textwidth]{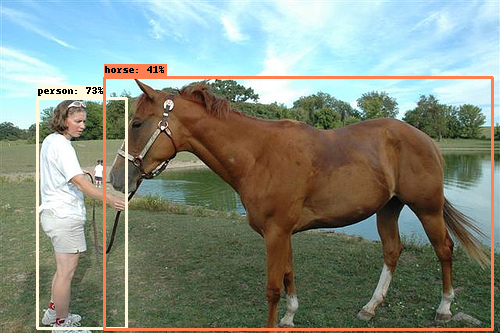}} 
	\subfigure{\includegraphics[width=0.23\textwidth]{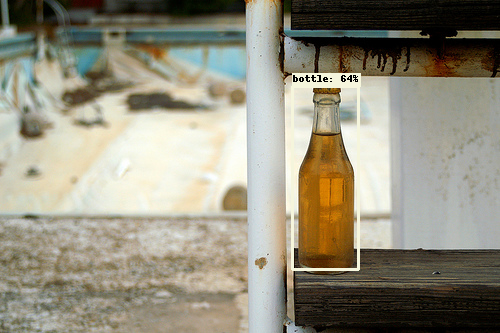}} \\
	\subfigure{\includegraphics[width=0.23\textwidth]{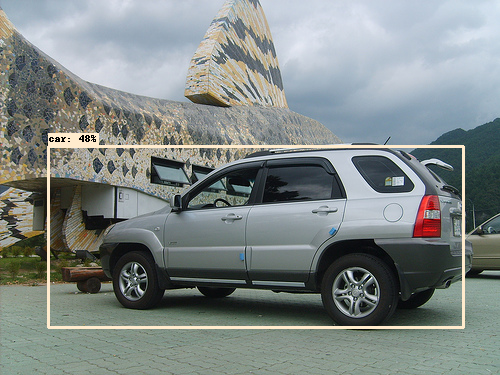}} 
	\subfigure{\includegraphics[width=0.23\textwidth]{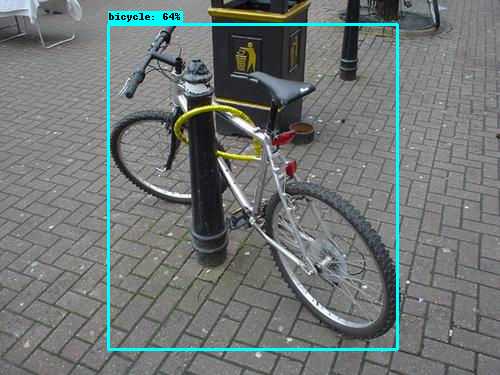}} \\
	\caption{Object detection results on PASCAL VOC dataset.} \label{object result}
	\vspace{0.2in}
\end{figure}

Fig. \ref{map} shows the object detection performance of SNN-YOLOv3 as the number of training epochs increases. In the figure, the green and red curves represent the performance trends for the CMD method with $T = 4$ and $T = 6$, respectively, while the blue curves illustrate the accuracy variation for the rate decoding method with $T = 6$. From the experimental results, we observe the following conclusions: (1) In general, CMD brings higher performance compared to rate decoding. (2) As the current mean decoded SNN is trained with a larger number of time steps, its performance will further increase. The results clearly indicate that when utilizing rate decoding, the SNN network struggles to learn, leading to consistently low accuracy. In contrast, the CMD method shows a remarkable ability to improve accuracy with increasing number of training iterations. This illustrate the exceptional effectiveness of the CMD method for object detection tasks in SNNs.

Moreover, the remarkable performance of SNN-YOLOv3 is also shown in the other examples in Fig. \ref{object result}. The SNN-YOLOv3 precisely locates and classifies various object categories within images, including person, cars, and bicycles which proves its excellent object localisation capability.

\subsection{SNN-YOLOv3 energy efficiency}
In order to assess the outstanding energy efficiency of SNN-YOLOv3, we compare the computational operations of SNN-YOLOv3 and YOLOv3 within the realm of digital signal processing. Within convolutional deep neural networks, the convolutional layer is the main computational region, where the multiply-accumulate (MAC) operation is the main executive operation. 
However, the operation performed in the spiking neural network is an accumulation (AC) operation because spiking events are binary events. The input current is integrated or accumulated into the membrane potential only when the neuron received a spike. For a fair comparison, we focus on the number of MACs and ACs consumed during single-image object detection. According to the literature \cite{horowitz20141}, 32-bit floating-point MAC operations consume 4.6 pJ and 32-bit floating-point AC operations consume 0.9 pJ.

\begin{figure}[tbp] 
	\centering 
	\includegraphics[width=0.45\textwidth]{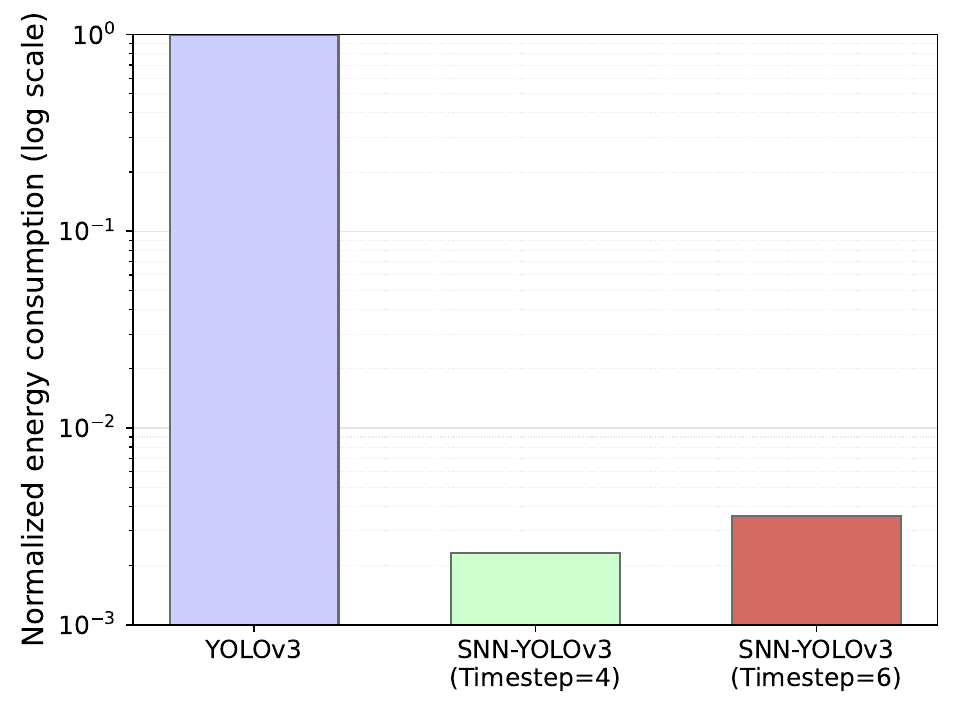} 
	\caption{The normalized energy comparison of YOLOv3 and SNN-YOLOv3 for MAC and AC operations.} \label{Energy}
\end{figure}

Based on these operation energy results, we calculated the energy consumption of YOLOv3 and SNN-YOLOv3 by multiplying the FLOPs (floating point operations) and the energy consumption per MAC or AC operation. if it is a SNN model, it needs to be further multiplied by time step. According our simulations, the FLOPs for ANN-YOLOv3 and SNN-YOLOv3 were $66.19$ and $0.425$ GFLOPs, respectively. Fig. \ref{Energy} shows the results, where SNN-YOLOv3 is more than 158 times energy efficient than YOLOv3 in 32-bit FL operations both under the $T=4$ and $T=6$.

\subsection{Comparison with the State-of-the-Art}

\renewcommand{\arraystretch}{1.6} 
\setlength{\tabcolsep}{4.8pt} 
\begin{table}[]
	\caption{Performance and energy consumption comparison between the proposed method and previous work on PASCAL VOC dataset.}
	\vspace{0.15in}
		\begin{tabular}{cccccc}
			\toprule 
			\multicolumn{5}{c}{\textbf{SNN-YOLOv3 (Ours)}}                                                            \\ \hline
			Method  & mAP(\%)        & Time step   & \multicolumn{1}{c|}{FLOPs}    & Energy            \\ \hline
			SG+CMD  & \textbf{61.87} & \textbf{6} & \multicolumn{1}{c|}{4.25E+08} & \textbf{6.38E-06} \\ \hline
			\multicolumn{5}{c}{\textbf{Spiking-YOLO \cite{kim2020spiking}}}                                                          \\ \hline
			Method  & mAP(\%)        & Time step   & \multicolumn{1}{c|}{FLOPs}    & Energy            \\ \hline
			ANN-SNN & 51.83          & 3500       & \multicolumn{1}{c|}{4.90E+07} & 4.29E-04          \\ 
			\bottomrule 
		\end{tabular}
	\label{comparison}
\end{table}

We compare our approach with other state-of-the-art ANN-to-SNN conversion methods on the PASCAL VOC dataset to achieve SNN object detection on non-trivial datasets. Our calculation result as shown in Table \ref{comparison}.  We calculated the energy consumption of SNN-YOLOv3 running on a neuromorphic chip (TrueNorth) and compared it to Spiking YOLO. the GFLOPS/W of TrueNorth is 300 GFLOPs/W, and we defined a time step as 1 ms (1 kHz synchronization signal in TrueNorth) \cite{merolla2014million}. For SNN-YOLOv3, ours proposed method can achieve 61.87\% mAP using only 6 time steps. Compare to Spiking-YOLO, we have nearly 10\% increase in mAP while requiring nearly two orders of magnitude less energy consumption. Considering that the TrueNorth chip was initially introduced in 2014, we can expect increased energy and computational efficiency as neuromorphic chips advance and produce better results.

\section{Conclusion}
In this paper, we introduce the energy-efficient SNN-YOLOv3, which is the first SNN model that implement object detection using surrogate gradient. It achieves the state-of-the-art performance (61.87\% mAP) on the PASCAL VOC dataset using only 6 time steps. Compared to previous work, we can accomplish object detection with less energy consumption. In addition, we proposed current mean decoding method for solving regression problem in SNN, which provides a different approach to work out more advanced machine learning problems with deep SNNs.

\bibliographystyle{IEEEbib}
\bibliography{strings, refs}

\begin{thebibliography}{10}

\bibitem{maass1997networks}
Wolfgang Maass,
\newblock ``Networks of spiking neurons: the third generation of neural network
  models,''
\newblock {\em Neural networks}, vol. 10, no. 9, pp. 1659--1671, 1997.

\bibitem{merolla2014million}
Paul~A Merolla, John~V Arthur, Rodrigo Alvarez-Icaza, Andrew~S Cassidy, Jun
  Sawada, Filipp Akopyan, Bryan~L Jackson, Nabil Imam, Chen Guo, Yutaka
  Nakamura, et~al.,
\newblock ``A million spiking-neuron integrated circuit with a scalable
  communication network and interface,''
\newblock {\em Science}, vol. 345, no. 6197, pp. 668--673, 2014.

\bibitem{pei2019towards}
Jing Pei, Lei Deng, Sen Song, Mingguo Zhao, Youhui Zhang, Shuang Wu, Guanrui
  Wang, Zhe Zou, Zhenzhi Wu, Wei He, et~al.,
\newblock ``Towards artificial general intelligence with hybrid tianjic chip
  architecture,''
\newblock {\em Nature}, vol. 572, no. 7767, pp. 106--111, 2019.

\bibitem{indiveri2015neuromorphic}
Giacomo Indiveri, Federico Corradi, and Ning Qiao,
\newblock ``Neuromorphic architectures for spiking deep neural networks,''
\newblock in {\em 2015 IEEE International Electron Devices Meeting (IEDM)}.
  IEEE, 2015, pp. 4--2.

\bibitem{mainen1995reliability}
Zachary~F Mainen and Terrence~J Sejnowski,
\newblock ``Reliability of spike timing in neocortical neurons,''
\newblock {\em Science}, vol. 268, no. 5216, pp. 1503--1506, 1995.

\bibitem{roy2019towards}
Kaushik Roy, Akhilesh Jaiswal, and Priyadarshini Panda,
\newblock ``Towards spike-based machine intelligence with neuromorphic
  computing,''
\newblock {\em Nature}, vol. 575, no. 7784, pp. 607--617, 2019.

\bibitem{lee2016training}
Jun~Haeng Lee, Tobi Delbruck, and Michael Pfeiffer,
\newblock ``Training deep spiking neural networks using backpropagation,''
\newblock {\em Frontiers in neuroscience}, vol. 10, pp. 508, 2016.

\bibitem{shrestha2018slayer}
Sumit~B Shrestha and Garrick Orchard,
\newblock ``Slayer: Spike layer error reassignment in time,''
\newblock {\em Advances in neural information processing systems}, vol. 31,
  2018.

\bibitem{diehl2015fast}
Peter~U Diehl, Daniel Neil, Jonathan Binas, Matthew Cook, Shih-Chii Liu, and
  Michael Pfeiffer,
\newblock ``Fast-classifying, high-accuracy spiking deep networks through
  weight and threshold balancing,''
\newblock in {\em 2015 International joint conference on neural networks
  (IJCNN)}. ieee, 2015, pp. 1--8.

\bibitem{sengupta2019going}
Abhronil Sengupta, Yuting Ye, Robert Wang, Chiao Liu, and Kaushik Roy,
\newblock ``Going deeper in spiking neural networks: Vgg and residual
  architectures,''
\newblock {\em Frontiers in neuroscience}, vol. 13, pp. 95, 2019.

\bibitem{rueckauer2017conversion}
Bodo Rueckauer, Iulia-Alexandra Lungu, Yuhuang Hu, Michael Pfeiffer, and
  Shih-Chii Liu,
\newblock ``Conversion of continuous-valued deep networks to efficient
  event-driven networks for image classification,''
\newblock {\em Frontiers in neuroscience}, vol. 11, pp. 682, 2017.

\bibitem{kim2020spiking}
Seijoon Kim, Seongsik Park, Byunggook Na, and Sungroh Yoon,
\newblock ``Spiking-yolo: spiking neural network for energy-efficient object
  detection,''
\newblock in {\em Proceedings of the AAAI conference on artificial
  intelligence}, 2020, vol.~34, pp. 11270--11277.

\bibitem{neftci2019surrogate}
Emre~O Neftci, Hesham Mostafa, and Friedemann Zenke,
\newblock ``Surrogate gradient learning in spiking neural networks: Bringing
  the power of gradient-based optimization to spiking neural networks,''
\newblock {\em IEEE Signal Processing Magazine}, vol. 36, no. 6, pp. 51--63,
  2019.

\bibitem{wu2018spatio}
Yujie Wu, Lei Deng, Guoqi Li, Jun Zhu, and Luping Shi,
\newblock ``Spatio-temporal backpropagation for training high-performance
  spiking neural networks,''
\newblock {\em Frontiers in neuroscience}, vol. 12, pp. 331, 2018.

\bibitem{kheradpisheh2020temporal}
Saeed~Reza Kheradpisheh and Timoth{\'e}e Masquelier,
\newblock ``Temporal backpropagation for spiking neural networks with one spike
  per neuron,''
\newblock {\em International Journal of Neural Systems}, vol. 30, no. 06, pp.
  2050027, 2020.

\bibitem{zhou2021temporal}
Shibo Zhou, Xiaohua Li, Ying Chen, Sanjeev~T Chandrasekaran, and Arindam
  Sanyal,
\newblock ``Temporal-coded deep spiking neural network with easy training and
  robust performance,''
\newblock in {\em Proceedings of the AAAI conference on artificial
  intelligence}, 2021, vol.~35, pp. 11143--11151.

\bibitem{cao2015spiking}
Yongqiang Cao, Yang Chen, and Deepak Khosla,
\newblock ``Spiking deep convolutional neural networks for energy-efficient
  object recognition,''
\newblock {\em International Journal of Computer Vision}, vol. 113, pp. 54--66,
  2015.

\bibitem{han2020rmp}
Bing Han, Gopalakrishnan Srinivasan, and Kaushik Roy,
\newblock ``Rmp-snn: Residual membrane potential neuron for enabling deeper
  high-accuracy and low-latency spiking neural network,''
\newblock in {\em Proceedings of the IEEE/CVF conference on computer vision and
  pattern recognition}, 2020, pp. 13558--13567.

\bibitem{auge2021survey}
Daniel Auge, Julian Hille, Etienne Mueller, and Alois Knoll,
\newblock ``A survey of encoding techniques for signal processing in spiking
  neural networks,''
\newblock {\em Neural Processing Letters}, vol. 53, no. 6, pp. 4693--4710,
  2021.

\bibitem{diehl2015unsupervised}
Peter~U Diehl and Matthew Cook,
\newblock ``Unsupervised learning of digit recognition using
  spike-timing-dependent plasticity,''
\newblock {\em Frontiers in computational neuroscience}, vol. 9, pp. 99, 2015.

\bibitem{redmon2018yolov3}
Joseph Redmon and Ali Farhadi,
\newblock ``Yolov3: An incremental improvement,''
\newblock {\em arXiv:1804.02767}, 2018.

\bibitem{everingham2010pascal}
Mark Everingham, Luc Van~Gool, Christopher~KI Williams, John Winn, and Andrew
  Zisserman,
\newblock ``The pascal visual object classes (voc) challenge,''
\newblock {\em International journal of computer vision}, vol. 88, pp.
  303--338, 2010.

\bibitem{horowitz20141}
Mark Horowitz,
\newblock ``1.1 computing's energy problem (and what we can do about it),''
\newblock in {\em 2014 IEEE international solid-state circuits conference
  digest of technical papers (ISSCC)}. IEEE, 2014, pp. 10--14.

\end{thebibliography}

\end{document}